\documentclass[10pt,twocolumn,letterpaper]{article}

\usepackage[pagenumbers]{cvpr} 

\newcommand{\arch}[0]{U-Shape Mamba}

\usepackage{subcaption}
\usepackage{algorithm}
\usepackage{algpseudocode}
\usepackage{adjustbox}
\usepackage{afterpage}

\definecolor{cvprblue}{rgb}{0.21,0.49,0.74}
\usepackage[pagebackref,breaklinks,colorlinks,allcolors=cvprblue]{hyperref}

\def\paperID{18} 
\def\confName{CVPR}
\def\confYear{2025}

\title{\arch{}: State Space Model for faster diffusion}

\author{Alex Ergasti$^1$\qquad Filippo Botti$^1$ \qquad Tomaso Fontanini$^1$ \\ Claudio Ferrari$^2$ \qquad Massimo Bertozzi$^1$ \qquad Andrea Prati$^1$ \\
$^1$ University of Parma, Department of Engineering and Architecture. Parma, Italy\\
$^2$ University of Siena, Department of Information engineering and mathematics. Siena, Italy \\
{\tt\small \{alex.ergasti, filippo botti, tomaso.fontanini\}@unipr.it}\\\tt\small{claudio.ferrari@unisi.it, \{massimo.bertozzi, andrea.prati\}@unipr.it} 
}

\begin{document}
\maketitle
\begin{abstract}
Diffusion models have become the most popular approach for high-quality image generation, but their high computational cost still remains a significant challenge. To address this problem, we propose U-Shape Mamba (USM), a novel diffusion model that leverages Mamba-based layers within a U-Net-like hierarchical structure. By progressively reducing sequence length in the encoder and restoring it in the decoder through Mamba blocks, USM significantly lowers computational overhead while maintaining strong generative capabilities. Experimental results against Zigma, which is currently the most efficient Mamba-based diffusion model, demonstrate that USM achieves one-third the GFlops, requires less memory and is faster, while outperforming Zigma in image quality. Frechet Inception Distance (FID) is improved by 15.3, 0.84 and 2.7 points on AFHQ, CelebAHQ and COCO datasets, respectively. These findings highlight USM as a highly efficient and scalable solution for diffusion-based generative models, making high-quality image synthesis more accessible to the research community while reducing computational costs.
\end{abstract}    
\section{Introduction}

Diffusion models today represent the state of the art for image generation. In particular, many existing approaches take inspiration from Latent Diffusion Model (LDM) \cite{rombach2022high} which employs an encoder-decoder structure and a U-Net backbone in order to efficiently execute the diffusion process in the latent space. This allows to greatly reduce the computational resources necessary for the training and inference process. Then, in order to further boost the scalability of these models, the U-Net backbone was substituted with a transformer-based one in Dit \cite{peebles2023scalable} and U-ViT \cite{bao2023all}. Nevertheless, the hardware requirements for these models is still huge and their widespread success raises questions in terms of finding sustainable solutions to make them available to a larger portion of the research community while also reducing their carbon footprint when many inference operations needs to be executed.

Recently, State Space Models (SSMs) \cite{gu2021efficiently,gu2021combining} became popular due to their efficiency. Among them,  Mamba \cite{gu2023mamba} was proposed as a novel solution for long sequence modeling. Mamba allows to reach results comparable to the ones of transformers but requiring much less resources. In particular, its complexity scales linearly with the sequence length rather than quadratically, allowing to drastically reduce its memory requirements. Even if Mamba was originally designed to work with 1D sequences, it was soon adapted to 2D sequence modeling reaching impressive results in classic computer vision tasks \cite{liu2024vmamba} such as image classification, object detection and semantic segmentation. 

Exploiting the efficiency of Mamba and the image generation capability of transformer-based diffusion models, Zigma \cite{hu2024zigma} proposed an architecture capable of generating samples with the same quality of DiT or U-ViT, but with half the GFlops. Additionally, authors also proposed a novel scan paradigm to enforce better spatial continuity.

In this paper we propose an additional step towards lighter and efficient generative models introducing \arch{} (USM). USM further reduces the computational requirements of Zigma while improving its generation quality. More in detail, we propose a novel backbone for diffusion models that is similar to the original U-Net concept, but it is composed by a series of Mamba-based blocks in place of convolutional layers. In each block, the sequence dimension is gradually reduced up until a bottleneck after which the original dimension is restored. Skip connections are added between blocks in order to avoid losing important information during this process. Ultimately, the contributions of this paper are the following:
\begin{itemize}
    \item A novel Mamba-based diffusion model called \arch{} (USM) which greatly reduces the hardware requirements for training and evaluation. The proposed system requires less memory, is faster, and has $1/3$ the GFlops w.r.t. Zigma while still generating samples with a higher quality. The model can also be conditioned with textual inputs, enhancing its flexibility;
    \item  A U-Net-like backbone with Mamba-based blocks in which, after each encoder block, the sequence length is reduced by a factor of 4, while, after each decoder block, the sequence length is increased up until the original dimension. Corresponding blocks in the encoder and decoder are linked via skip connections.
\end{itemize}
\section{Related Work}

\noindent\textbf{Diffusion Models: } In the past years Diffusion Models saw widespread success and became the standard solution for generative tasks \cite{rombach2022high,blattmann2023stable,peebles2023scalable,bao2023all, ergasti2024controllablefacesynthesissemantic}. To generate new samples, diffusion models need to learn a backward process that gradually removes noise from the input. This procedure requires several steps and hinders their efficiency and speed. Recently, Flow Matching was proposed \cite{lipman2023flow,liu2022flow} allowing to learn a straight path between noise and data and thus making the diffusion process much simpler mathematically, obtaining as second effect a faster diffusion process. In our work, we utilize Flow Matching to train the proposed architecture and pair it with a Mamba-based backbone in order to limit the computation resources and increase the inference speed. This allows our system to work even on limited hardware which was not possible with previous solutions.

\noindent\textbf{Mamba: } Transformers have been widely and successfully used in computer vision tasks \cite{dosovitskiy2020image}. Nevertheless, their complexity increases quadratically with respect to the number of patches in the image and this could hinder their use in scenarios where limited computational resources are available or the hardware cost needs to be constrained. To tackle this limitation, State Space Models (SSM) such as \cite{gu2021efficiently,gu2021combining} represent a viable solution due to their linear complexity. Among these, Mamba \cite{gu2023mamba} was recently proposed. Mamba works by making the SSM parameters input-dependent and was proved to be the best SSM model. After its introduction, Mamba was applied to deep learning especially to solve computer vision tasks \cite{liu2024vmamba,guo2024mambair,Botti_2025_WACV}. Furthermore, Zigma \cite{hu2024zigma} proposed a Mamba-based diffusion model backbone greatly reducing the computational costs. Finally, our system USM further improves this aspect allowing to reduce the computational burden even more while boosting the quality of the generated results at the same time.

\section{Architecture}

\begin{figure*}[t!]
    \centering
    \includegraphics[width=\textwidth]{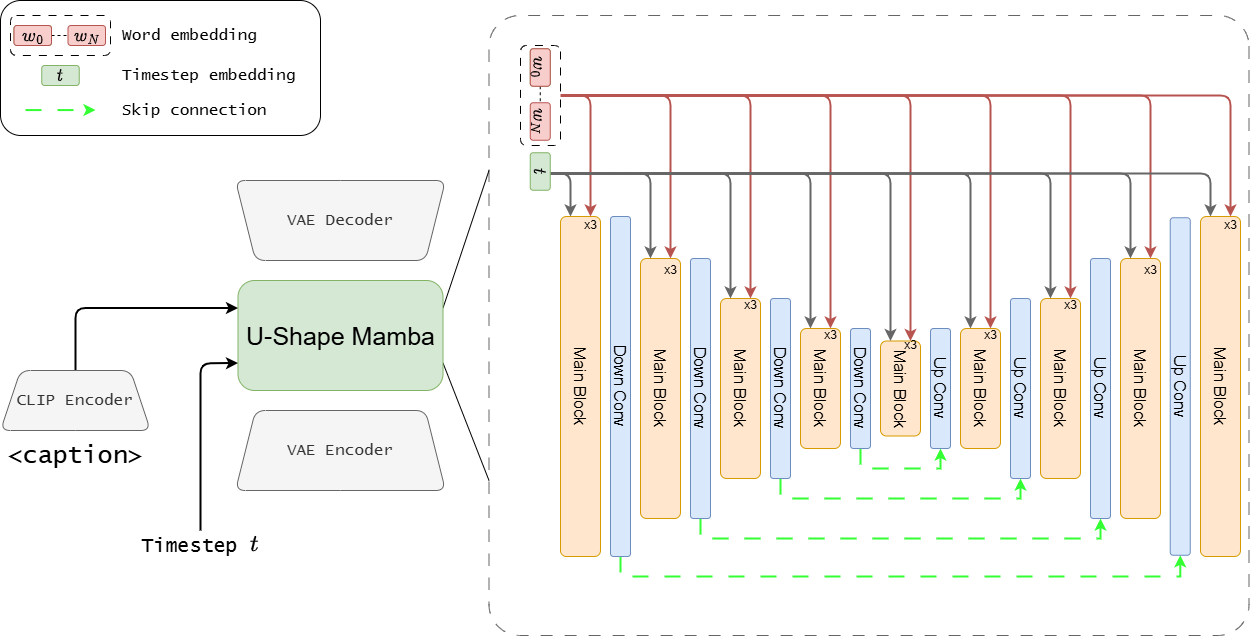}
    \caption{An overview of the architecture of our models. The input image $x$ is first processed by a VAE encoder to reduce dimensionality, producing a lower-resolution representation. This is then reshaped and projected to obtain a feature map $\phi$, which passes through U-Shape Mamba core blocks. The encoder progressively reduces dimensions through downsampling convolutional layers every three blocks. This results in a highly compressed feature map $\phi_{\text{middle}}$ in the final bottleneck Mamba-based block. Then, the decoder mirrors the encoder, using transposed convolutions to progressively restore the spatial resolution. Skip connections between the encoder and the decoder are added to avoid losing important information during the encoding phase.}
    \label{fig:arch}
\end{figure*}
\begin{figure*}[t!]
    \centering
    \begin{subfigure}{0.37\textwidth}
        \includegraphics[width=\textwidth]{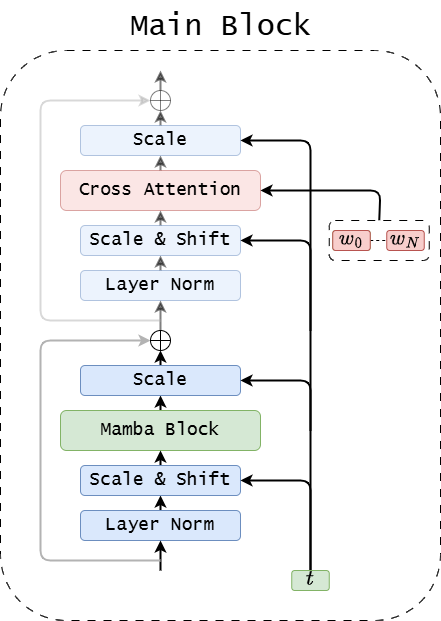}
        \caption{}
        \label{fig:main-block}
    \end{subfigure}
    \hfill
    \begin{subfigure}{0.337\textwidth}
        \includegraphics[width=\textwidth]{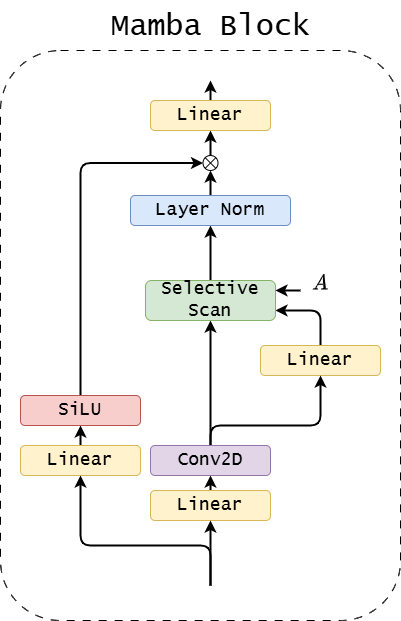}
        \caption{}
        \label{fig:mamba-block}
    \end{subfigure}
    \hfill
    \begin{subfigure}{0.272\textwidth}
        \includegraphics[width=\textwidth]{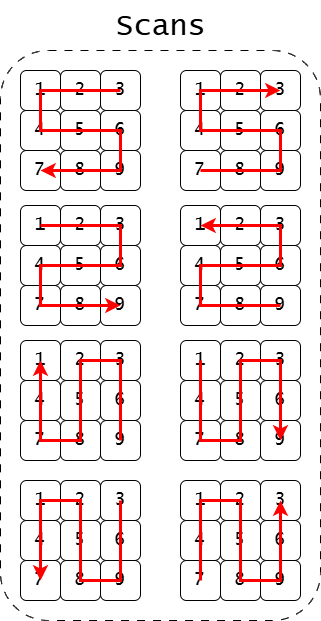}
        \caption{}
        \label{fig:scans}
    \end{subfigure}
    \caption{a) The main block of our architecture. b) Mamba block. c) All the scans used inside the selective scan block. }
\end{figure*}

\subsection{Mamba}
Mamba is a sequence-to-sequence model, derived from the SSMs \cite{gu2023mamba,dao2024transformers}, which learns to map one sequence $x(t) \in \mathbb{R}$ to another $y(t) \in \mathbb{R}$ by using an internal state $h(t) \in \mathbb{R}^N$, where $N$ is the hidden dimension. The inner equations that describe Mamba, and SSMs, follow the ordinary differentiable equation (ODE) linear system: 
\begin{equation}
     \begin{split}
     h'(t) &= A h(t) + B x(t)  \\        
     y(t) &= C h(t) + D x(t) 
     \end{split}
     \label{eq:mamba}
\end{equation}
\noindent where the matrices $A \in \mathbb{R}^{N\times N},B\in \mathbb{R}^{N\times 1},C\in \mathbb{R}^{1\times N} $ and $ D\in \mathbb{R}$ are learnable. Since the equations are continuous in time, in order to be used in a deep learning architecture, a discretization phase is required. Here, following \cite{gu2023mamba}, we applied the zero-order holder (ZOH) rule, where $\Delta$ represents the step parameter and $\Bar{A},\Bar{B}$ represent the discretized matrices. The system in a RNN form becomes:
\begin{equation}
     \begin{split}
     h_{k} &= \Bar{A}h_{k-1} + \bar{B}x_k  \\        
     y_{k} &= Ch_k + Dx_k
     \end{split}
     \label{eq:mamba_disc}
\end{equation}
Finally, the most important intuition is that in Mamba, differently from other SSMs, the matrices are both learnable and input-dependent. This means that, given an input $x$, the matrices $B, C$ and $\Delta$ are obtained with a linear fully-connected layer as in the following:
\begin{equation}
     \begin{split}
     B = \mathrm{Lin_B}(x), C = \mathrm{Lin_C}(x), \Delta = \mathrm{Lin_{\Delta}}(x)
     \end{split}
     \label{eq:mamba_input_dependency}
\end{equation}

\subsection{Rectified Flow}
Diffusion models work by gradually adding noise to a data distribution (forward process) and then learning to remove that noise to get back the original distribution (backward process).
Indeed, a generic forward process which maps a sample $x$ from the data distribution to gaussian noise $\epsilon$ can be expressed as:
\begin{equation}
    z = \alpha x + \beta\epsilon
\end{equation}
\noindent Flow Matching \cite{liu2022flow,lipman2023flow}  is a specific parametrization of that forward process which aims to connect the noise distribution and the data distribution in the simplest way, using a linear interpolation. Thus, we can substitute $\alpha=(t)$ and $\beta=(1-t)$ to obtain a linear interpolation which depends on a time $t$:
\begin{equation}
    z(t) = tx + (1-t)\epsilon\quad t\in[0,1]
    \label{eq:forward}
\end{equation}

The neural network then has to learn the backward process, which can be solved using an ODE. Thus, the model learns $v_\theta(z(t),t)$ which is an estimation of the derivative of Eq.~\eqref{eq:forward}:
\begin{equation}
    v = \frac{d}{dt}z(t) = x-\epsilon
\end{equation}
Then, the final loss can be calculated as the mean squared error (MSE) between the real $v$ and the estimate $v_\theta$ as follow:
\begin{equation}
    \mathcal{L}(\theta) =  w(t) \| v_\theta(z(t),t) - (x-\epsilon) \|^2
\end{equation}

\noindent where $w(t)$ represents a weighting function that prioritizes certain regions of the trajectory. Empirical evidence \cite{esser2024scaling} suggests that using a logit-normal distribution for $w(t)$ with zero mean and unit variance yields superior performance. The training algorithm is explained in Alg. \ref{alg:training}.

During inference, the backward process is calculated using the Euler approximation method, illustrated in Alg. \ref{alg:inference}. By using Flow Matching it is possible to greatly reduce the number of steps required to generate a sample w.r.t. to vanilla diffusion models.

\begin{algorithm}[h!]
\caption{Training Algorithm}
\begin{algorithmic}[1]
\Require{Dataset $\mathcal{D}$, neural network $v_\theta$ with parameters $\theta$, batch size $B$, learning rate $\alpha$, number of iterations $N$}
\State Initialize model parameters $\theta$ randomly
\For{$i = 1$ to $N$}
    \State $\mathbf{x} \sim \mathcal{D}$
    \State $\mathbf{\epsilon} \sim \mathcal{N}(0, I)$
    \State $\mathbf{t} \sim \mathcal{U}(0, 1)$
    \State $\mathbf{z}(\mathbf{t}) \gets \mathbf{t}\mathbf{x} + (1-\mathbf{t})\mathbf{\epsilon}$
    \State $\mathbf{v} \gets \mathbf{x} - \mathbf{\epsilon}$
    \State $\hat{v} \gets v_\theta(\mathbf{z}(\mathbf{t}), \mathbf{t})$
    \State $\mathbf{w} \gets \text{logit-normal}(\mathbf{t}; 0, 1)$
    \State $\mathcal{L} \gets \frac{1}{B}\sum_{i=0}^{B-1}\mathbf{w}_i\cdot\mathrm{MSE}(\hat{v}_i-v_i)$
    \State $\theta \gets \theta - \alpha \cdot \nabla_\theta \mathcal{L}$
\EndFor
\State \Return trained model parameters $\theta$
\end{algorithmic}
\label{alg:training}
\end{algorithm}

\begin{algorithm}[h!]
\caption{Inference Euler Algorithm}
\begin{algorithmic}[1]
\Require{Neural network $v_\theta$ with parameters $\theta$, number of inference steps $T$}
\State $dt = 1/T$
\State $\mathbf{z}_{T} \sim \mathcal{N}(0,I)$
\State $\mathbf{t}=1$
\For{$i = T$ to $1$}
    \State $\hat{v} \gets v_\theta(\mathbf{z}, \mathbf{t})$
    \State $\mathbf{z}_{i-1} \gets \mathbf{z}_{i}+d\mathbf{t}\cdot \hat{v}$
    \State $\mathbf{t} \gets \mathbf{t}-d\mathbf{t}$
\EndFor
\State \Return $\mathbf{z}_0$
\end{algorithmic}
\label{alg:inference}
\end{algorithm}

\subsection{U-Shape Mamba}
Our architecture, called \arch{} (USM), is a  Mamba-based encoder-decoder network designed for efficient diffusion process (Fig.~\ref{fig:arch}). The architecture is modular and is composed by a series of main blocks. In particular, we decided to utilize 25 blocks strategically distributed across three components: 12 blocks in the encoder, a single block serving as the bottleneck layer, and 12 blocks in the decoder.

Given an image $x\in\mathbb{R}^{C\times H\times W}$, Firstly a VAE encoder (the same as in LDM \cite{rombach2022high}) is used to reduce its dimensionality, obtaining an output in $\mathbb{R}^{c\times h\times w}$, where $h=H/f$, $w=W/f$, with $f=8$ and $c=4$. Then, the input is reshaped and projected to obtain a sequence $\phi\in\mathbb{R}^{h\cdot w\times D}$, where $D$ is the hidden size. This is equivalent of using a patch size of 1x1. The sequence $\phi$ then passes through USM main blocks.

The encoder features a progressive dimensionality reduction scheme. After every three blocks, we implement a downsampling operation. Given the sequence $\phi \in \mathbb{R}^{h \cdot w \times D}$, this is reshaped to have a shape of $h \times w \times D$ and  passed through a convolutional layer with kernel 2 and stride 2 obtaining a new feature of dimension $h/2 \times w/2 \times D$. After a final reshaping the new sequence becomes $\hat{\phi} \in \mathbb{R}^{\frac{h \cdot w}{4} \times D}$. This allows to greatly reduce the computational resources required by the model. 

After the encoder, we obtain a sequence $\phi_\mathrm{middle}\in\mathbb{R}^{\frac{h \cdot w}{64}\times D}$. Then, a single block serves as the bottleneck between the encoder and decoder. Finally, the decoder mirrors the encoder's structure but reverses the dimensionality changes. Before each group of three blocks, we employ transposed convolutions (Up Conv) that progressively restore spatial resolution, effectively countering the encoder downsampling operations.

To mitigate information loss during the encoding-decoding process, we implement dense skip connections between corresponding encoder and decoder layers. The skip tensors ($\phi_{enc}$) from the encoder are concatenated with the decoder features ($\phi_{dec}$) along the hidden dimension, then linearly projected in order to reduce the hidden dimension by a factor of 2, since the concatenated vector has a double hidden dimension size. These connections enable the network to leverage both high-level semantic information and fine-grained spatial details, resulting in precise diffusion.

\noindent\textbf{Main Block Structure.} Both the encoder, middle and decoder blocks share the same configuration, illustrated in Fig.~\ref{fig:main-block}. More in detail, firstly, an AdaLN layer \cite{peebles2023scalable} is employed in order to scale and shift the sequence based on the timestep $t$, then the core Mamba Block, illustrated in Fig.~\ref{fig:mamba-block}, is placed to model the sequence. Additionally,  an optional cross attention module can be added to the main block after the Mamba portion of it. The role of this additional component is to handle the case when text condition $\mathbf{w} = \{w_o\cdots,w_N\}$ is used during training and inference. 

\begin{table*}
    \centering
    \begin{tabular}{c|ccc|ccc}
        \toprule
        & \multicolumn{3}{c|}{w/o text} & \multicolumn{3}{c}{w/ text}\\
        \midrule
         Model & GFlops & Memory (MB) & Time (ms) & GFlops & Memory (MB) & Time (ms)   \\
         \midrule
         Zigma  & 64.12 & 1281.37 & 97.34 & 105.94 & 1619.28 & 118.46 \\
         Ours   & \textbf{20.66} & \textbf{790.67} & \textbf{17.52} & \textbf{40.84} & \textbf{1136.09} & \textbf{22.92} \\
         \bottomrule
    \end{tabular}
    \caption{Comparison of performances between our model and zigma with and without text condition.}
    \label{tab:computational_eff}
\end{table*}

\noindent\textbf{Scans Configuration.} Instead of opting for a fixed scans configuration in each block, we follow the same approach of Zigma \cite{hu2024zigma}. In USM each Mamba block utilizes a different scan configuration, as illustrated in Fig.~\ref{fig:scans}. The architecture cycles through all 8 possible scan configurations in a deterministic sequence. Specifically, the first block employs the first scan configuration, the second block uses the second configuration, and so on up until the eighth block adopting the eighth configuration. This pattern then repeats, with the ninth block returning to the first scan configuration. This cyclic pattern continues throughout all 25 blocks in the architecture, ensuring diverse directional information processing across the network.

\section{Results}

In this section the results of the proposed model will be analyzed with a particular focus on the computational resources required during the inference phase. Our objective is to prove that USM can reach better generation quality w.r.t. Zigma while requiring a fraction of the processing power.

\begin{figure*}[]
    \centering
    \includegraphics[width=\textwidth]{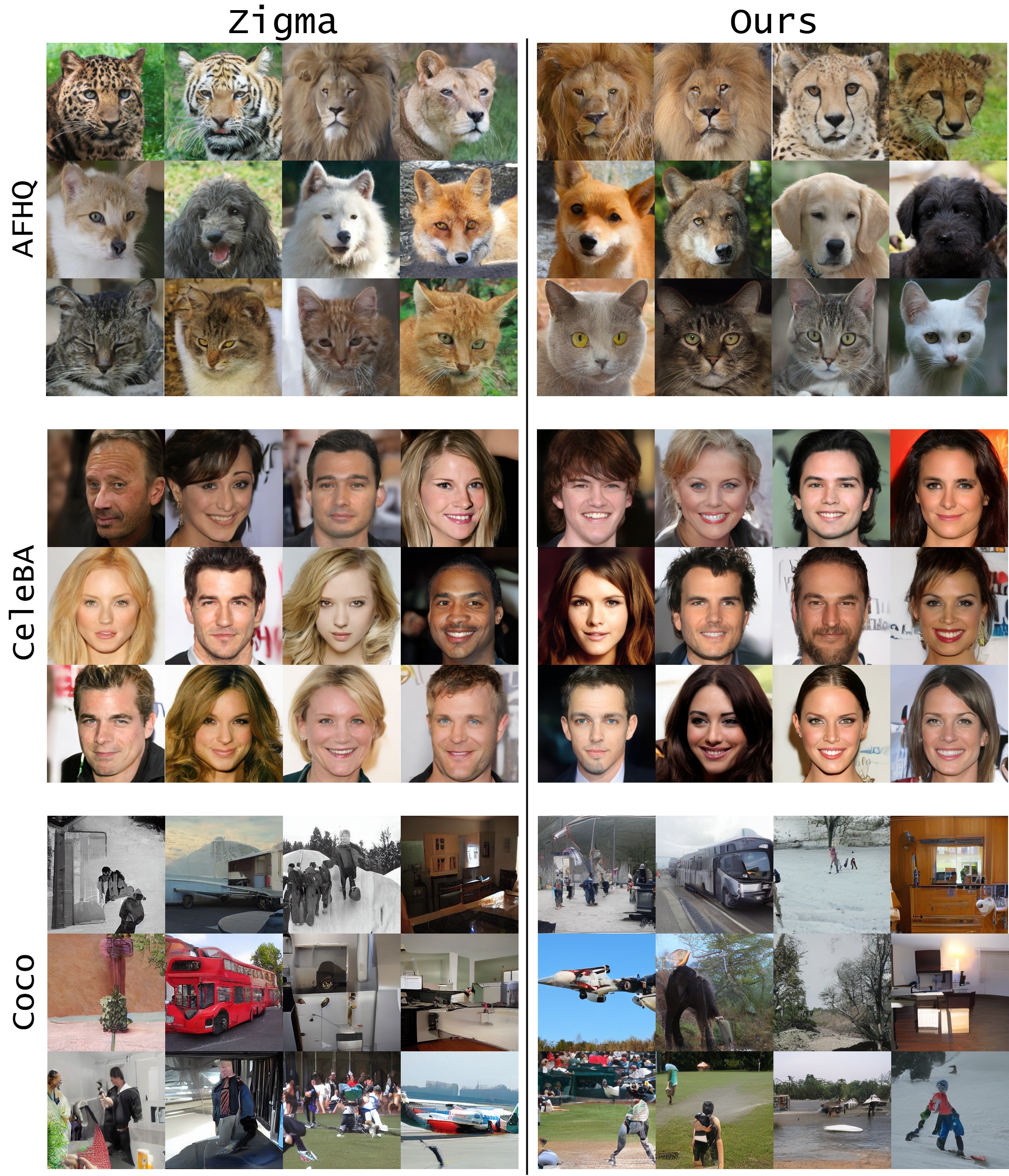}
    \caption{Comparisons between results obtained with Zigma and with our USM model on three different datasets: AFHQ, CelebA and COCO.}
    \label{fig:examples}
\end{figure*}

\subsection{Datasets}
We trained and tested our model on three datasets commonly employed in generative tasks:
\begin{itemize}
    \item Animal FacesHQ (AFHQ) \cite{choi2020stargan} is composed by 15000 images of high quality animal faces at 512 $\times$ 512 resolution;
    \item CelebAHQ \cite{karras2017progressive} contains 30000 face images of celebrities at 1024 $\times$ 1024 resolution;
    \item COCO \cite{lin2014microsoft} is an object detection dataset of more than 300k images. Since images are paired with a textual description, we used it to perform text-to-image generation.
\end{itemize}

\subsection{Training Details}

Our model was trained using images of size 256 $\times$ 256 for 400k iterations on a single NVIDIA L40S, using Adam as optimizer with a learning rate of 1e-4 and a batch size of 8. All the samples are then generated using 25 steps. We retrained ZigMa for AFHQ and CelebA for the same number of iterations, the same learning rate and batch size. Regarding COCO we used the pretrained checkpoint provided by ZigMa authors.

\subsection{Evaluation Metrics}
To evaluate the quality of the generated samples we employ the Frechet Inception Distance (FID) metric \cite{heusel2017gans}, which is a standard metric in generative models. FID measures the distance between the distribution of the real images and the distribution of the generated images.

Additionally, since the objective of this work is to propose an efficient generative model with a Mamba-based backbone, we also evaluate our architecture in terms of speed, memory occupancy and GFlops. Regarding speed, we decided to measure the average time spent to perform a single diffusion step when generating the samples. All test are performed using a batch size of 1.

\subsection{Quantitative Evaluation}
Regarding the quantitative evaluation of the proposed system, we decided to compare it with Zigma \cite{hu2024zigma}. Indeed, Zigma can be seen as the baseline model which we improved by proposing a more efficient backbone. Our proposed USM model significantly improves computational efficiency and image generation quality compared to Zigma. 

Regarding the computational efficiency, results can be seen in Table \ref{tab:computational_eff}. Thanks to its design, USM achieves a remarkable reduction in GFlops both when the text condition is utilized and when it is not. More in detail, USM consumes only 20.66 GFlops during inference while Zigma requires 64.12 GFlops without text condition. On the other side, when text is utilized during inference, our system consumes 40.84 GFlops w.r.t the 105.94 of Zigma. 

Additionally, the memory requirements is far less in the proposed model w.r.t. Zigma: 39\% less memory is required without text condition and 30\% less memory is required with text condition. Finally, our system is much faster than Zigma. In fact, USM is 73\% faster that Zigma without text condition and 81\% faster with text condition.

\begin{table}[h!]
    \centering
    \begin{tabular}{c|ccc}
            \toprule
             & \multicolumn{3}{c}{FID $\downarrow$} \\
            \midrule
            Model & AFHQ & CelebA & CoCo  \\
            \midrule
            Zigma & 32.17 & 22.54 & 41.80 \\
            Ours  & \textbf{16.87} & \textbf{21.80} & \textbf{39.10}  \\
            \bottomrule
    \end{tabular}
    \caption{Comparison between generated examples between our model and zigma on AFHQ, CelebA and CoCo.}
    \label{tab:res-notext}

\end{table}

In Table \ref{tab:res-notext} a quantitative evaluation on the quality of the generated results on all the datasets is presented. Indeed, USM obtains a lower FID score (16.87, 21.80 and 39.10, respectively) than Zigma (32.17, 22.54 and 41.80, respectively), indicating superior image quality even if it requires less computational requirements. 

These results confirm that USM provides a more efficient and effective alternative to state-of-the-art diffusion models, making high-quality image generation more accessible with reduced hardware constraints. Indeed, our model can be deployed on a broader range of devices due to its efficiency. It can also assure a lower power consumption making the inference process cheaper and with reduced carbon footprint.

\subsection{Qualitative Evaluation}

In Fig.~\ref{fig:examples} we report a visual comparison of results generated by Zigma and our model. Even with much fewer computational requirements, our model can generate high quality samples in each of the three datasets, proving the efficiency of its design.

\subsection{Ablation study}

\begin{figure}
    \centering
    \includegraphics[width=\linewidth]{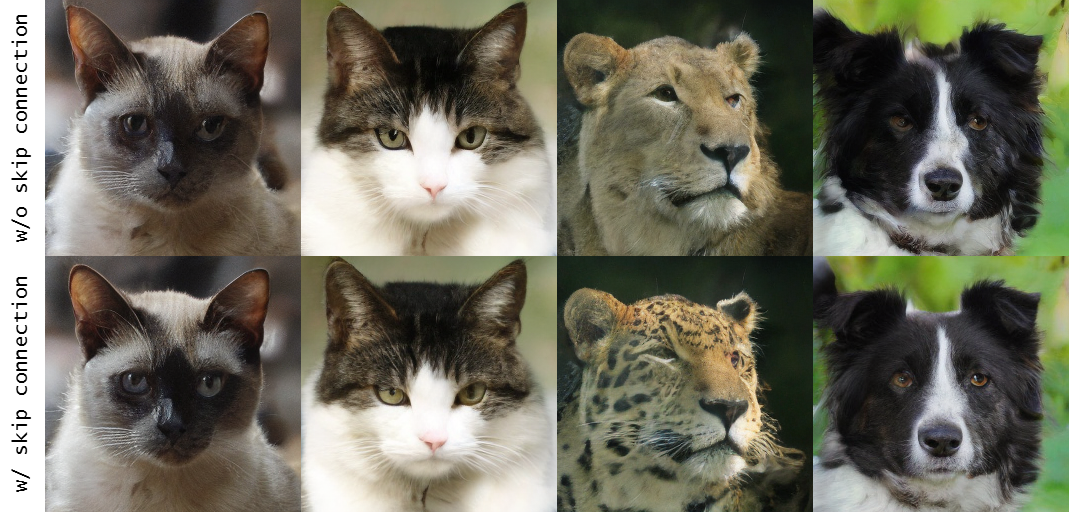}
    \caption{Comparisons on AFHQ between our model without and with skip connections.}
    \label{fig:abl}
\end{figure}

\begin{table}
    \centering
    \begin{center}
    \adjustbox{width=\linewidth}{
        \begin{tabular}{c|ccc|c}
            \toprule
            \multicolumn{5}{c}{AFHQ} \\
            \midrule
            Model & GFlops & Memory (MB) & Time (ms) & FID $\downarrow$\\
            \midrule
            Ours w/o skip conn &  7.98 & 736.63 & 17.02 & 22.84 \\
            Ours w/ skip conn  & 20.66 & 790.67 & 17.52 & 16.87 \\
            \bottomrule
        \end{tabular}}
        \caption{Ablation study which compare our model with and without skip connection.}
        \label{tab:ablation}
    \end{center}
\end{table}

In Table~\ref{tab:ablation} the importance of skip connections in USM is highlighted. Removing them increases the FID score from 16.87 to 22.84, reinforcing their role in preserving essential information during the diffusion process. On the other side, the GFlops without skip connections are greatly reduced due to the fact that no linear projection is needed since there is no feature concatenation between the encoder and the decoder.

Finally, in Fig.~\ref{fig:abl} a qualitative comparisons between samples generated without skip connections and with skip connections is presented. The animal faces generated with the proposed system equipped with skip connection are sharper and more detailed. Nevertheless, results without skip connections are still visually appealing and a final user could also consider using this version of the model in scenarios where the hardware limitations are exceptional.

\section{Conclusion}
In this work, we introduced \arch{} (USM), a novel Mamba-based diffusion model designed to significantly reduce computational costs while maintaining high-quality image generation. By leveraging a U-Net-like hierarchical structure with Mamba blocks and skip connections, USM achieves state-of-the-art efficiency, requiring only one-third of the GFlops compared to Zigma while producing superior results, as demonstrated by lower FID scores across multiple datasets. Our experiments confirm that USM is not only computationally efficient but also highly effective in generating high-quality images, even when incorporating text conditioning. The ablation study further underscores the importance of skip connections in preserving critical information. With its reduced hardware requirements, USM makes advanced diffusion-based generative models more accessible to a broader research community, paving the way for more sustainable and scalable AI-driven image generation. 
\section{Acknowledgements}
This work was funded by ``Partenariato FAIR (Future Artificial Intelligence Research) - PE00000013, CUP J33C22002830006" funded by the European Union - NextGenerationEU through the italian MUR within NRRP, project DL-MIG.

{
    \small
    \bibliographystyle{ieeenat_fullname}
    \bibliography{main}
}


\end{document}